\documentclass{article}

\usepackage{arxiv}

\usepackage[utf8]{inputenc} 
\usepackage[T1]{fontenc}    
\usepackage{hyperref}       
\usepackage{url}            
\usepackage{booktabs}       
\usepackage{amsfonts}       
\usepackage{nicefrac}       
\usepackage{microtype}      
\usepackage{lipsum}
\usepackage{graphicx}
\graphicspath{ {./images/} }

\usepackage{amsmath}
\usepackage{hyperref} 
\usepackage[nodisplayskipstretch]{setspace}
\usepackage{fontawesome}

\usepackage[linesnumbered,ruled,vlined]{algorithm2e}

\usepackage[ruled,vlined]{algorithm2e}  

\SetCommentSty{mycommfont}

\SetKwInput{KwInput}{Input}                
\SetKwInput{KwOutput}{Output} 

\usepackage[compact]{titlesec}         
\AtBeginDocument{
\setlength\abovedisplayskip{0pt}
\setlength\belowdisplayskip{0pt}}
\usepackage{multirow} 
  
\title{CL3: A Collaborative Learning Framework for the Medical Data Ensuring Data Privacy in the Hyperconnected Environment}

\author{Mohamamd Zavid Parvez \and
Rafiqul Islam \and
Md Zahidul Islam \and \\
School of Computing, Mathematics and Engineering, Charles Sturt University, NSW, Australia\and\\
}

\begin{document}
\maketitle
\begin{abstract}
In a hyperconnected environment, medical institutions are particularly concerned with data privacy when sharing and transmitting sensitive patient information due to the risk of data breaches, where malicious actors could intercept sensitive information. A collaborative learning framework, including transfer, federated, and incremental learning, can generate efficient, secure, and scalable models while requiring less computation, maintaining patient data privacy, and ensuring an up-to-date model. This study aims to address the detection of COVID-19 using chest X-ray images through a proposed collaborative learning framework called CL3. Initially, transfer learning is employed, leveraging knowledge from a pre-trained model as the starting global model. Local models from different medical institutes are then integrated, and a new global model is constructed to adapt to any data drift observed in the local models. Additionally, incremental learning is considered, allowing continuous adaptation to new medical data without forgetting previously learned information. Experimental results demonstrate that the CL3 framework achieved a global accuracy of 89.99\% when using Xception with a batch size of 16 after being trained for six federated communication rounds. A demo of the CL3 framework is available at \url{https://github.com/zavidparvez/CL3-Collaborative-Approach} to ensure reproducibility.
\end{abstract}


\section{Introduction}
Preventing the privacy threats of medical data in a hyperconnected environment is crucial for maintaining patient trust and complying with regulations when extensive network devices are interconnected and communicate with each other in real-time~\cite{hyperconnected}. The healthcare sector is one of the most affected areas by security and privacy issues due to the increasing number of medical devices, which are diverse and complex in their technological aspects~\cite{healthcare}. Malicious actors are targeting hospitals to gain access to their systems. As a result, we frequently see news reports about cyber-attacks in hospitals~\cite{securityHospitals}. The report indicates that the healthcare sector experienced the highest rate of data breaches among all industries, accounting for 22\% of breaches in Australia during the last six months of 2023. In comparison, financial institutions faced only 10\% of data breaches~\cite{securityHospitals}. 

In a hyperconnected environment, medical data (e.g., medical images) is continuously generated and often shared with stakeholders~\cite{medicalDataShare}. Therefore, the sequence of upcoming medical data needs to be considered for training machine learning models. These models can remain up-to-date with current medical data without starting from scratch to address their adaptability and memory efficiency. Additionally, machine learning models often need to be generated with an emphasis on reducing training time, improving performance with limited data, and enhancing learning from subsets~\cite{goodModel}.

To address issues such as privacy, improve performance with less time and limited data, and keep models up-to-date with recent medical information, we can consider a collaborative learning framework such as federated learning, transfer learning, and incremental learning. For the experiment with the collaborative learning framework, we use chest X-ray images from COVID-19 and healthy subjects, as chest X-ray images are widely used for medical research and detection purposes based on machine learning strategies~\cite{coroDet}, \cite{covid1},\cite{covid2}.

Several research studies have been conducted for COVID-19 detection using Convolutional Neural Network (CNN) models, including \cite{coroDet} proposed a 22-layer CNN model for COVID-19 detection and achieved 91.2\% accuracy for 2-class classification. Hussein et al.\cite{covid2} used a deep CNN and achieved 98.55\% accuracy for 2-class classification. Moreover, Gupta et al.~\cite{covid1} considered an AI-based approach for prediction of COVID-19.

Some research works have considered Federated Learning for COVID-19 screening including~\cite{covidFL}, \cite{covidFL1}, \cite{covidFL2}. Dasha et al. used the Federated Learning framework for screening COVID-19 using X-ray images and achieved 96.5\% accuracy, 96.8\% sensitivity, and 96.2\% specificity using ResNet50 with 5-fold cross-validation~\cite{covidFL}. Chowdhury et al. also used a Federated learning framework and achieved a global accuracy of 99.59\% after 3-federated communication rounds \cite{covidFL1}. Kandati et al.~\cite{covidFL2} considered the Federated Learning approach based on particle swarm optimization and achieved 96.15\% accuracy using the COVID-19 infection database.

Incremental learning (i.e., online learning or continual learning) is a machine learning approach where models are updated continuously as new data becomes available and models are updated. Very little research has been done considering incremental learning for COVID-19 detection using X-ray database. Malik et al. used an incremental learning-based cascaded model and detected tuberculosis from X-ray images with 83.32\% accuracy~\cite{covidIL}.

Different deep learning approaches, such as transfer learning, federated learning, and incremental learning, have distinct purposes, including transferring knowledge, ensuring privacy, and generating up-to-date models. Although numerous research studies have focused on COVID-19 detection using X-ray images, they have typically considered only one or two of these learning approaches~\cite{coroDet}, \cite{covid1}, \cite{covid2}, \cite{covidFL}, \cite{covidFL1}, \cite{covidFL2}, \cite{covidIL}. Therefore, we have combined all three learning approaches and proposed a collaborative learning framework called \emph{CL3}, ensuring knowledge transfer, privacy, and the latest model for COVID-19 detection using chest X-ray images. To the best of our knowledge, we are the first to propose this type of collaborative learning framework for COVID-19 detection.

The key contributions in this paper are as follows:
\begin{itemize}
\item We propose a novel collaborative learning framework, referred to as \emph{CL3}, based on transfer learning, federated learning, and incremental learning which effectively detects COVID-19. 

\item  We conduct experiments to evaluate \emph{CL3} framework using a gold standard dataset with ground truth which reveals the effectiveness of \emph{CL3} in comparison with existing pre-trained models with different batch sizes.
\end{itemize}

The rest of this paper is organized as follows: Section~\ref{proposed_Method} introduces a novel collaborative learning framework called \emph{CL3}. Section~\ref{exp_setup_Eval} outlines the experimental setup and evaluation, while Section~\ref{PerformanceAnalysis} discusses the performance analysis. Finally, Section~\ref{conclusion} provides the conclusion.

\section{Our Proposed Method: A Collaborative Learning Framework (CL3)}\label{proposed_Method}
\begin{figure*}[hbt!]
   \centering
   \includegraphics[width=1.05\linewidth]{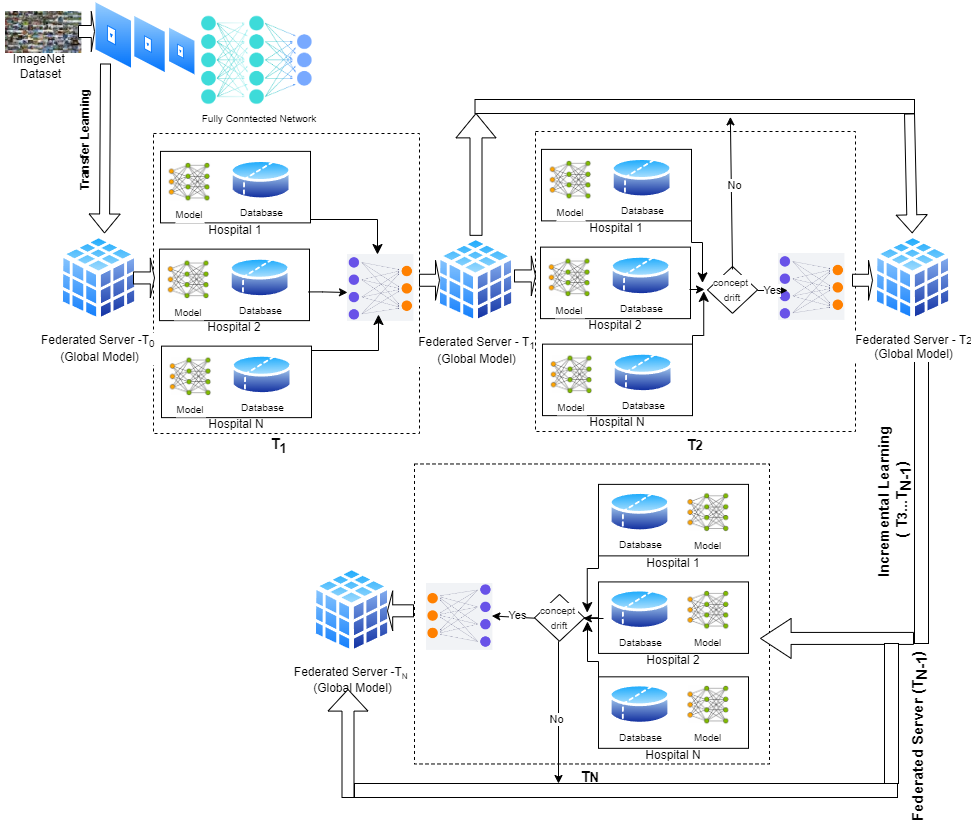}
   \caption{Block diagram of the proposed Collaborative Learning (i.e., CL3) Framework.}
   \label{CL3}
 \end{figure*}

A Collaborative Learning Framework, called CL3, comprises three learning approaches: Transfer Learning, Federated Learning, and Incremental Learning, illustrated in Fig. \ref{CL3} and Algorithm 1, and further discussed with background knowledge in the following subsections.

\subsection{Transfer Learning with ImageNet}\label{tl_ImageNet}

ImageNet is a large visual dataset containing over 14 million images annotated into 21K classes~\cite{imageNet}. In the proposed method, we apply a pre-trained deep convolutional neural network architecture (DCNNA), specifically Extreme Inception (Xception), on the ImageNet dataset to transfer knowledge (i.e., transfer learning approach) and achieve an initial global model (see Fig. \ref{CL3}) for faster convergence on training data. We use Xception, implemented using Python libraries such as Keras and TensorFlow~\cite{coroNet} because it can reduce parameters and computational cost~\cite{xception}.

\subsection{Federated Learning}\label{fl}
Federated Learning~\cite{federatedLearning} generates robust models using decentralized data from various hospitals without centralizing the data, thereby mitigating privacy concerns and adhering to regulations such as the General Data Protection Regulation (GDPR) and the Health Insurance Portability and Accountability Act (HIPAA)\cite{regulation}. This learning approach involves two types of servers: a central server and local servers\cite{federatedLearningMoinul}. The central server contains a global model, while the local servers contain hospital data. Multiple hospitals train their models locally without sharing their raw data but share their model updates with the central server. The central server aggregates the local servers' updates to build a global model by considering the previous global model weights which is illustrated in Fig. \ref{CL3}. In this way, hospitals keep their data safe and secure, ensuring patient data privacy.

\subsection{Incremental Learning}\label{il}
Incremental learning is an adaptive approach that can accommodate new data instances as they arrive without requiring all historical data. It also updates its parameters incrementally without forgetting previously learned knowledge~\cite{incrementalLearning}. This learning approach efficiently uses memory and storage by addressing model drift caused by changes in the underlying data on the local server over time.

\section{Experimental Setup and Evaluation}\label{exp_setup_Eval}

All experiments are conducted on Windows 11 Home, 64-bit, with an 11th Gen Intel(R) Core(TM) i5-1135G7 2.42 GHz processor and 16.0 GB RAM.

Algorithm 1 outlines the steps of CL3, which considers the ImageNet dataset and X-ray images as inputs and provides $G_{model}$ and $L_{model}$ as outputs. The initial global model is created using the ImageNet dataset and the Xception pre-trained model by leaving off the fully connected head. This head is replaced with a sequential CNN model featuring a customized classification head, including a max pooling layer, a flattened layer, and a dense fully connected layer.

For the local models, each representing a different hospital, X-ray images are used with a batch size of 16 for training. The local models are trained for at least 25 epochs to ensure convergence. If model drift occurs, the locally trained weights are sent to the global model. The global model weights are then updated based on the weights received from the local models. This process is repeated for 6 communication rounds to achieve satisfactory performance for the global model.

\begin{algorithm}[H]
\SetAlgoLined
\LinesNumbered

\KwIn{ImageNet\_Data: ImageNet Dataset \\ 
      \quad\quad\quad Xray\_images: X-ray Images}
\KwResult{G\_model: Global Model \\ 
          \quad\quad\quad L\_model: Local Model}
           
\SetKwFunction{FTransfer}{TransferLearningModule}
\SetKwFunction{FFederated}{FederatedLearningModule}
\SetKwFunction{FIncremental}{IncrementalLearningModule}
\SetKwProg{Fn}{Function}{:}{}
   
\Fn{\FTransfer{ImageNet\_Data}}{
    Transfer\_knowledge $\gets$ \text{Xception}(ImageNet\_Data)\;
    \KwRet Transfer\_knowledge\;
}

\Fn{\FIncremental{G\_model}}{
    \For{$i = 1$ \KwTo $N$}{
        G\_model $\gets$ \FFederated{G\_model}\;
    }
    \KwRet G\_model\;
}

\Fn{\FFederated{G\_model}}{
    \If{G\_model $\neq$ NULL}{
        \For{$i = 1$ \KwTo Hospital\_Count}{
            \If{L\_model == Model\_Drift}{
                Aggregated $\gets$ L\_model(Xray\_images)\;
            }
        }
    }
    G\_model $\gets$ Aggregated\;
    \KwRet G\_model\;
}

\caption{CL3 Framework}
\label{CL3Framework}
\end{algorithm}

The steps of the algorithm (see Algorithm 1: CL3 Framework), which is better aligned with Fig \ref{CL3}, are as follows:

\begin{enumerate}
\item Build an initial global model using transfer learning from the ImageNet pre-trained model.

\item Each hospital will use the initial model to build its first local model, combined with all other hospital models using Federated Learning. Federated Learning shares the combined model with each hospital again so that they can train the model again based on their local data. This process continues multiple times until the models converge.

\item When a new batch of data arrives, incremental learning is used for each hospital separately to first test whether there is a concept drift for each hospital. If a concept drift is detected, the model for that hospital gets updated using incremental learning. The updated model is then shared for Federated Learning again.

\item Repeat the process whenever a new batch of data arrives for a hospital.
\end{enumerate}

The code is available in the GitHub repository~\cite{CodeCL3}, and the dataset can be obtained from the CovRecker repository~\cite{dataset},\cite{CovReckerDataset}.


\subsection{Dataset}
\begin{figure*}[hbt!]
   \centering
   \includegraphics[width=0.9\linewidth]{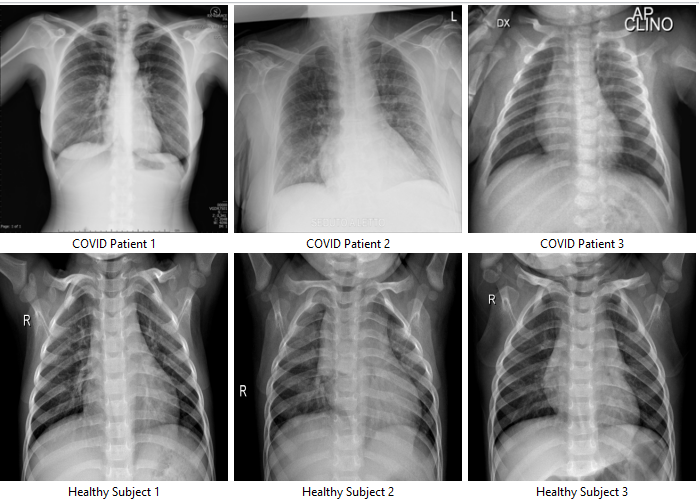}
   \caption{X-ray images, where the top three are images of COVID-19 patients, and the bottom three are images of healthy subjects.}
   \label{x-rayImages}
 \end{figure*}

We collected our experimental dataset from the CovRecker repository~\cite{dataset}. The dataset consists of 2,330 X-ray images, with 1,178 images from healthy subjects, 1,122 images from COVID-19 patients used for model generation, and the remaining 30 images reserved for testing the model. We split our training dataset into 5 groups (namely: Hospital 1, Hospital 2, and so on) to apply the Federated Learning approach. The detailed dataset distribution is illustrated in Table \ref{covidDataset}. For the training phase, we consider 6 increments of X-ray images for each hospital for result analysis. Fig. \ref{x-rayImages} demonstrated the X-ray images of COVID-19 patients and healthy subjects.

\begin{table*}[h]
\centering
\caption{Dataset characteristics} 
\begin{tabular}{|c|cc|}
\hline
\multirow{2}{*}{\textbf{No. of Hospital}} & \multicolumn{2}{c|}{\textbf{X-ray Images}}                                 \\ \cline{2-3} 
                                   & \multicolumn{1}{l|}{\textbf{Healthy Subjects}} & \multicolumn{1}{l|}{\textbf{COVID-19 Patients}} \\ \hline
1                                  & \multicolumn{1}{c|}{216}             & 237                                 \\ \hline
2                                  & \multicolumn{1}{c|}{220}             & 213                                 \\ \hline
3                                  & \multicolumn{1}{c|}{258}             & 198                                 \\ \hline
4                                  & \multicolumn{1}{c|}{265}             & 237                                 \\ \hline
5                                  & \multicolumn{1}{c|}{219}             & 237                                 \\ \hline
\end{tabular}
\label{covidDataset}
\end{table*}

\section{Performance Analysis}
\label{PerformanceAnalysis}
We can consider several metrics to justify the model performance, such as classification accuracy ($CA$), precision ($PRE$), recall ($REC$), and $F1 Score$. From an effective model, we expect to find more matches (higher REC) with fewer false positives (higher PRE). The $F1 Score$ indicates an overall performance range from 0 to 1. Moreover, we can consider classification accuracy to assess the model's performance. The formulae for $CA$, $PRE$, $REC$, and $F1 Score$ are as follows:

 \begin{equation}
 CA=\frac{TP+TN}{TP+FP+TN+FN} \label{eq}
 \end{equation}

 \begin{equation}
 PRE=\frac{TP}{TP+FP} \label{eq}
 \end{equation}

 \begin{equation}
 REC=\frac{TP}{TP+FN} \label{eq}
 \end{equation}

 \begin{equation}
 F1 Score=\frac{2*PRE*REC}{PRE+REC} \label{eq}
 \end{equation}

\begin{table*}[hbt!]
\centering
\caption{Results based on Transfer Learning approach using Five Hospital X-ray iamges} 
\begin{tabular}{|c|c|c|c|c|}
\hline
\textbf{Hoptial} & \textbf{CA (\%)} & \textbf{F1 Score (0...1)} & \textbf{PRE (\%)} & \textbf{REC (\%)} \\ \hline
\textbf{1}       & 99.6             & 0.99                & 99.6              & 99.6              \\ \hline
\textbf{2}       & 99.8             & 0.99                & 99.8              & 99.8              \\ \hline
\textbf{3}       & 99.8             & 0.98                & 98.0              & 98.0              \\ \hline
\textbf{4}       & 99.4             & 0.99                & 99.4              & 99.4              \\ \hline
\textbf{5}       & 98.2             & 0.98                & 98.2              & 98.2              \\ \hline
\end{tabular}
\label{CNNModel}
\end{table*}

\begin{figure*}
\centering
\includegraphics[width=0.48\linewidth]{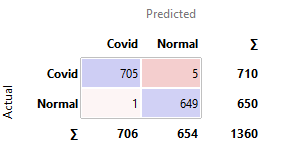}
\hspace{.005\linewidth} 
\includegraphics[width=0.48\linewidth]{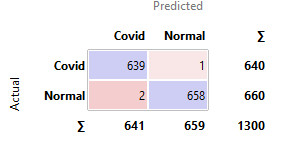}
\\[.5\baselineskip]
\includegraphics[width=0.48\linewidth]{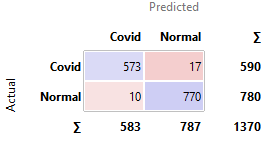}
\hspace{.005\linewidth} 
\includegraphics[width=0.48\linewidth]{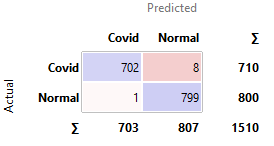}
\\[.5\baselineskip]
\includegraphics[width=0.48\linewidth]{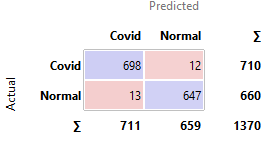}
\caption{The confusion matrix based on transfer learning with the pre-trained Inception model is organized as follows: the top-left represents Hospital 1, the top-right represents Hospital 2, the middle-left represents Hospital 3, the middle-right represents Hospital 4, and the bottom represents Hospital 5.}
\label{confusionMatrix}
\end{figure*}

In Table \ref{CNNModel}, we investigate the results using transfer learning based on an Inception pre-trained model and a fully connected network with 5 hidden layers, each containing 100 neurons. Various parameter values are explored, including the ReLU activation function and Adam optimization with a regularization value of $1*e^{-4}$, and random sampling with repeated train/test splits 10 times, using a training size of 70\%. The results indicate that the model performed well across all five hospital datasets. To justify the results from Table \ref{CNNModel}, we presented the confusion matrix in Fig. \ref{confusionMatrix} using all five hospital datasets. The model using X-ray images from Hospital 1, Hospital 2, and Hospital 4 performed slightly better compared to the other two hospitals, which aligns with the results in Table \ref{CNNModel}. Notably, the number of instances increased after generating image embeddings. In conclusion, Table \ref{CNNModel} shows that the single learning approach consistently outperforms the combined approach for every training data (see Fig. \ref{confusionMatrix}). However, while this method yields strong performance, it does not ensure data privacy or does not provide an up-to-date model.


\begin{table*}[hbt!]
\centering
\caption{Communication round accuracy with six Increments} 
\begin{tabular}{|c|c|c|c|}
\hline
\textbf{\begin{tabular}[c]{@{}c@{}}Increment \\ No.\end{tabular}} & \textbf{\begin{tabular}[c]{@{}c@{}}Accuracy (\%)\\ VGG16\\ (Batch Size=16)\end{tabular}} & \textbf{\begin{tabular}[c]{@{}c@{}}Accuracy (\%)\\ Xception\\  (Batch Size=8)\end{tabular}} & \textbf{\begin{tabular}[c]{@{}c@{}}Accuracy (\%)\\ Xception \\ (Batch Size=16)\end{tabular}} \\ \hline
1                                                                 & 76.66                                                                                    & 69.99                                                                                       & 89.99                                                                                        \\ \hline
2                                                                 & 83.33                                                                                    & 86.66                                                                                       & 89.99                                                                                        \\ \hline
3                                                                 & 60.00                                                                                    & 89.99                                                                                       & 89.99                                                                                        \\ \hline
4                                                                 & 80.00                                                                                    & 50.00                                                                                       & 86.66                                                                                        \\ \hline
5                                                                 & 83.33                                                                                    & 86.66                                                                                       & 89.99                                                                                        \\ \hline
6                                                                 & 83.33                                                                                    & 89.99                                                                                       & 89.99                                                                                        \\ \hline
\end{tabular}
\label{federatedResults}
\end{table*}


To address the issues of privacy and maintaining an up-to-date model, Table \ref{federatedResults} shows the performance accuracy of the proposed \emph{CL3} framework after six communication rounds with six increments, using two pre-trained models and two batch sizes. We compared the performance of VGG16 with a batch size of 16, Xception with a batch size of 8, and Xception with a batch size of 16. Among these, Xception with a batch size of 16 delivers the most consistent results across different increments and communication rounds. This consistency can be attributed to more stable gradients due to averaging over larger sample sizes, as well as the use of depthwise separable convolutions. Consequently, Xception with a batch size of 16 outperforms both VGG16 with a batch size of 16 and Xception with a batch size of 8.

\section{Conclusion} \label{conclusion}
Federated learning is an efficient approach to address privacy issues in hyperconnected environments. Additionally, pre-trained transfer learning is effective for reducing computational requirements and can generate a global model using a small dataset. To accommodate the need for up-to-date models with different time lags, we also consider incremental learning. By combining all three learning approaches, we propose a collaborative learning framework called \emph{CL3}, which addresses these issues and provides a more effective approach. We used X-ray images of COVID-19 patients and healthy subjects to validate our proposed framework. We compared the results using two pre-trained models, Xception and VGG16, with two batch sizes based on communication rounds six and six increments. Results demonstrated that Xception with a 16 batch size provides better results than the other pre-trained model and batch sizes because Xception uses depthwise separable convolutions to extract profound insights from the data.


\section{Acknowledgements}
The work has been supported by the Cyber Security Research Centre Limited whose activities are partially funded by the Australian Government’s Cooperative Research Centres Programme.


\end{document}